\documentclass[11pt]{article}
\usepackage{times}
\usepackage{url}
\usepackage{latexsym}
\usepackage{spconf,amsmath,amssymb,graphicx}
\usepackage[babel=true]{csquotes} 
\usepackage{algorithm}
\usepackage[noend]{algpseudocode}
\usepackage{flushend}
\usepackage{framed,color}
\usepackage{pgfplots}
\usepackage{subfig}

\title{Joint On-line Learning of a Zero-shot Spoken Semantic Parser and a Reinforcement Learning Dialogue Manager}

\name{Matthieu Riou\quad Bassam Jabaian\quad St\'ephane Huet\quad Fabrice Lef\`evre\thanks{This work has been partially carried out within the Labex BLRI (ANR-11-LABX-0036).}}
\address{
     CERI-LIA \\ University of Avignon \\ Avignon, France
}

\begin{document}
\maketitle
\begin{abstract}
Despite many recent advances for the design of dialogue systems, a true bottleneck remains the acquisition of data required to train its components. Unlike many other language processing applications, dialogue systems require interactions with users, therefore it is complex to develop them with pre-recorded data. Building on previous works, on-line learning is pursued here as a most convenient way to address the issue. Data collection, annotation and use in learning algorithms are performed in a single process. The main difficulties are then: to bootstrap an initial basic system, and to control the level of additional cost on the user side. Considering that well-performing solutions can be used directly off the shelf for speech recognition and synthesis, the study is focused on learning the spoken language understanding and dialogue management modules only. Several variants of joint learning are investigated and tested with user trials to confirm that the overall on-line learning can be obtained after only a few hundred training dialogues and can overstep an expert-based system.
\end{abstract}


\section{Introduction}
\label{intro}

While a new avenue of research on end-to-end deep-learning-based dialogue systems has shown promising results lately \cite{Serban2016, Wen2017}, a major hindrance remains the need of a huge quantity of data for these models to be trained efficiently. So far, in this case, it is not clear how can some initial (low cost) knowledge be leveraged for a warm start of the system development followed by on-line training with users as describe in~\cite{Ferreira13b, Gasic10}.

In the experiments reported here our underlying goal is to develop a system intended to be used in a neuroscience experiment. From inside a fMRI, users interact with a robotic platform, vocally powered by our system, which is live-recorded and displayed inside the head-antenna. It is then crucial to have it develop in French. Therefore not only it is not possible to use the publicly available corpora as the vast majority is in English~\cite{williams2013,Casanueva2017}. But also because a new task is targeted (see section~\ref{seq:exp}) for which no data is available yet. Likewise crowdsourcing to realise large-scale data collection is not affordable, as not platform (AMT or others) offers enough NLP-skilled French-speaking workers.

As a consequence in this work we still refer to a classical architecture, with proven capabilities, for goal-directed vocal interaction. It is basically a pipeline of modules dealing with the audio information from the user downstream; progressive treatments aims to first extract the content (speech recognition), then the meaning (semantic parsing, SP), to finally combine it with previous information (including grounding status) from the dialogue history (belief tracking) so that a policy can decide upon this dialogue state representation the next best action to perform according to some global criteria (generally dialogue length and success in reaching the goal). This step of dialogue management (DM) is then followed by processings to convey back the information upstream to the user: conversion of the dialogue manager action into natural language (NLG) followed by speech synthesis. The HIS architecture \cite{Young10} offers such a setup encompassing a global statistical framework to account for the relations between the data handled by the main modules of the system, allowing a reinforcement learning of the DM policy. In this system could have been implemented the most sample-efficient learning algorithms~\cite{Daubigney12}, from which on-line learning with direct interactions with the user could have been proposed~\cite{FerreiraCSL}. More recently on-line learning has been generalized to the input/output modules, SP and NLG, with protocols to control the cost of such operations during the system development (as in~\cite{ferreira2015, Ferreira2016, Riou2017}). In this work it is our first attempt to combine the on-line learning of SP and DM in a single phase of development. Not only it is expected to help speed-up and simplify the process, but also to benefit from intertwined improvements of the modules.

In dialogue systems, SP extracts a list of semantic concept hypotheses from an input sentence transcription of the user's query. This list is generally expressed as a sequence of Dialogue Acts (DAs) of the form \textit{acttype(slot=value)} and transmitted to the Dialogue Manager (DM) to make decision on the future action to perform.  State-of-the-art SPs are based on probabilistic approaches and trained with various machine learning methods to tag the user input with these semantic concepts~\cite{hahn2010, lefevre2007, deoras2013}. Dealing with supervised machine learning techniques requires a large amount of annotated data which are domain dependent and hardly available.

To deal with this limitation, Dauphin et al.~\cite{dauphin2014} proposed a zero-shot learning algorithm for Semantic Utterance Classification (SUC). This method tries to find a sentence-wise link between categories and utterances in a semantic space. A deep neural network can be trained on a large amount of non-annotated and unstructured data to learn this semantic space. In the same line, in~\cite{ferreira2015} was presented a zero-shot learning method for SP (ZSSP) based on word embeddings~\cite{mikolov2013}. This approach requires neither annotated data nor in-context data. Indeed, only the ontological description of the target domain and generic word embedding features (learned from freely available and general purpose data) are required to initiate the model. On top of that an active learning strategy based on an adversarial bandit has been proposed~\cite{Ferreira2016} in order to train ZSSP with a light and controlled supervision from the users.

In the same line of ideas, thanks to the sample-efficient RL algorithm KTD~\cite{Geist10}, an active learning scheme has also been proposed for the DM training which uses reward shaping~\cite{Ng99} to take into account local (turn-based) rewards from the user to offer a better control over the learning process and speed it up~\cite{FerreiraCSL}.

Since solutions exist for active on-line learning of both SP and DM sub-systems, we now consider their joint application to address the issue of the overall training of the system. First a direct application of existing  techniques is presented and tested. Both modules remain separated and the parameters of their on-line trainings are kept disjoint (a bandit algorithm for SP, a Q-learner for DM). Then a new possibility with shared parameters in a single Q-learner is also introduced and evaluated.

The reminder of this paper is organized as follows. After presenting the basis of the on-line learning versions of SP in Section~\ref{seq:zssp} and DM in Section~\ref{seq:dm}, we define the joint on-line learning strategies in Section~\ref{seq:joint}. Section~\ref{seq:exp} provides an experimental study with human evaluations of the proposed approaches. And we conclude in Section~\ref{seq:concl}.

\section{On-line Learning for Zero-shot SP}
\label{seq:zssp}

\begin{figure}[!t]
 \centering
 \begin{tabular}{c}
  \includegraphics[width=0.30\textwidth]{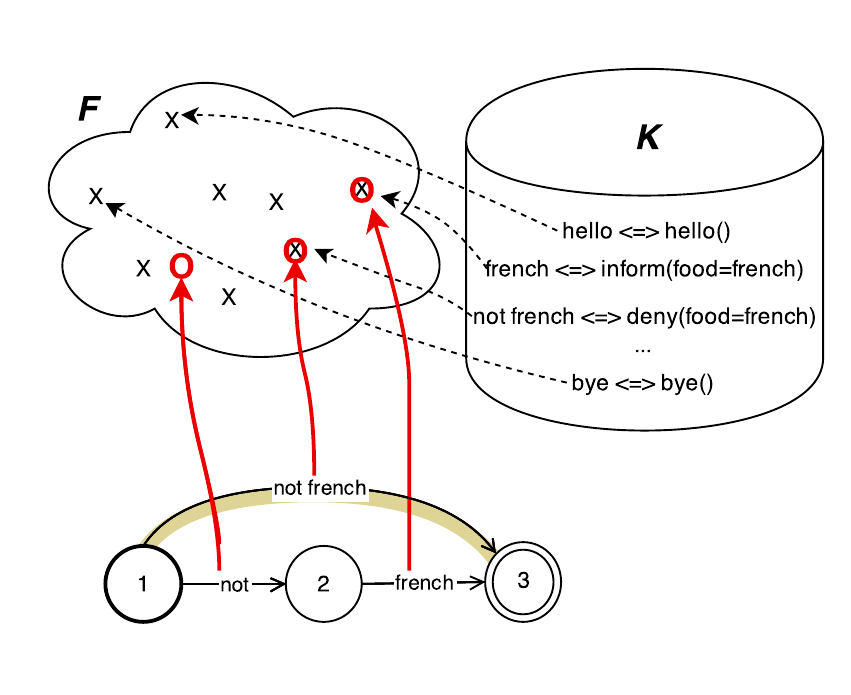}
 \end{tabular}
  \vspace{-10pt}
 \caption{Basics of the ZSSP Model}
 \label{fig:zssp}
\end{figure}

The SP model concerned by this study is the ZSSP model presented in~\cite{Ferreira2016} and illustrated in Fig~\ref{fig:zssp}. This latter makes use of a semantic knowledge base $K$ and a semantic feature space $F$. $K$ contains some examples of lexical chunks associated with each targeted DA and $F$ is a word embeddings learnt with neural network algorithms on large non-annotated open domain data~\cite{mikolov2013, bian2014}. The SP composes a scored graph of hypotheses from user utterances. All possible contiguous chunks are considered in the graph and a dot product between the k-most similar vectors and their corresponding assignment coefficients in the $K$ matrix is computed to attribute to each chunk a list of scored semantic hypotheses. A best-path decoding is performed in order to find the best semantic tags hypothesis for the considered user utterance.  

An on-line adaptation strategy is adopted, as presented in~\cite{Ferreira2016} and briefly recalled here. In this approach, at each dialogue iteration, the system chooses an adaptation action $i_t\in \mathcal{I}$ and uses the user feedback to update $K$.

The system gain $g(i_t)$, the user effort $\phi(i_t)$ and the loss function $l(i_t)$ for performing each action are defined and can be estimated during on-line training.   

Three possible actions are considered:
\begin{itemize}
    \item \textbf{Skip}: Skip the adaptation process for this turn ($\phi(\text{skip}) = 0$).
    \item \textbf{AskConfirm}: A yes/no question is presented to the user about the correctness of the selected DAs in the best semantic hypothesis. If the whole sentence is accepted, $\phi(\text{YesNoQuestions}) = 1$. Otherwise, $\phi(\text{YesNoQuestions})$ is equal to $1+$ the number of DA in the best semantic hypothesis (one yes/no confirmation request per DA).
    \item \textbf{AskAnnotation}: the user is asked to re-annotate the whole utterance. $\phi(\text{AskAnnotation}) = 1$ if the sentence is accepted straight away. Otherwise, the user will first inform the system about the word boundaries of the concept he plans to annotate ($+1$), and then the system will sequentially ask for $acttype$, $slot$ and $value$ if necessary ($+1$ per interim question) for each DA.
\end{itemize}

An adversarial bandit algorithm is used in order to find $i_1, i_2, \ldots,$ such that for every $t$, the system minimizes the loss $l (i_t)$. The loss function $l(i) \in [0,1]$ is calculated as follows:
$$l(i):=\underbrace{\gamma g(i)}_{\rm{system\ improvement}}+\underbrace{ (1-\gamma) \frac{\phi(i)}{\phi_{max}}}_{\rm{user\ effort}}, $$
where $\gamma\in[0,1]$ balances the importance of information improvement and user effort for the system and $\phi_{max}\in\mathbb{N}_+$ is the maximum number of exchanges between the system and the user (in a same turn/round). In this work, $\gamma$ has been set to $0.5$ for example.

\section{On-line Learning for RL Dialogue Manager}
\label{seq:dm}

The dialogue manager used in this paper adapts a system presented in~\cite{FerreiraCSL}. It is based on a POMDP-based dialogue management framework, the Hidden Information State (HIS) ~\cite{Young10}. In this setup, the system maintains a distribution over possible dialogue states (the belief state) and uses it to generate an adequate answer. A reinforcement learning (RL) algorithm is used to train the system by maximizing an expected cumulative discounted reward.

At each turn, the dialogue manager generates several possible answers, depending on its belief state. It generates 11 dialogue acts, matching the 11 summary acts described in Table~\ref{table:summary_acts}. Some can be deemed impossible at some point if no conversion to full action is possible (for instance \texttt{inform} if no entity is selected yet).

\begin{table}[t]
\center
\begin{tabular}{l l}
\hline
\textbf{Greet} & Greet user \\
\textbf{Bye} & End the dialogue \\ \hline
\textbf{BoldRQ} & Bold query request \\
\textbf{TentRQ} & Tentative query request \\
\textbf{Confirm} & Confirm an ungrounded piece of information \\
\textbf{FindAlt} & Find alternative database entity \\
\textbf{Split} & Distinguish two hypotheses \\
\textbf{Repeat} & Repeat \\
\textbf{Offer} & Offer a database entity \\
\textbf{Inform} & Give info about current offer \\
\textbf{QMore} & Query if the user wants more information \\
\end{tabular}
\caption{\label{table:summary_acts}Descriptions of the dialogue manager's summary acts}
\end{table}

The dialogue manager then chooses the best summary act according to the given context. To learn this policy, a RL approach is used: the KTDQ learning algorithm~\cite{geist2010managing}, derived from a Kalman-based Temporal Differences (KTD) framework. At each turn, the policy select a summary act to answer the user, then feedback is given by the users to score the response and update the policy. There are two types of feedback. The global feedback is given at the end of the dialogue by asking the user if the entire dialogue is a success or not. The social feedback $s_i$ is given at each turn $i$ to score the last response only. It is composed of two parts, the score given by the user to this last response (named additional-feedback $a_i$) minus a $\Psi$ function (to limit the influence of the additional-feedback to their turn), and the turn cost which allow to penalise too long dialogues by adding a negative score to each turn taken (named feedback $f_i$). 
$$s_i = f_i + (\theta a_i - \Psi) $$ \\
Here $\Psi =$ is the previous turn's additional-feedback $a_{i-1}$ and $\theta = 0.95$. At the end of the dialogue, the policy is updated according to all the collected feedbacks.

In this work, the global feedback value is set to 20 in case of success, 0 otherwise. The feedback $f_i$ is set to -1 for each turn and the additional-feedback $a_i \in\{-1,-0.5,0,0.5,1\}$.

\section{Joint On-line Learning}
\label{seq:joint}

In order to effectively learn on-line the dialogue system, the user needs to be able to both improve the SP model and the dialogue manager. Two different joint learning protocols are proposed to achieve it. 

The first one, referred to as \textbf{BR} hereafter, directly juxtaposes the bandit to learn the ZSSP and the Q-learner RL approach to learn the dialogue manager policy. An adversarial bandit algorithm as described in Section~\ref{seq:zssp} is applied for learning ZSSP and a Q-learner as described in Section~\ref{seq:dm} is used to learn the DM policy. The knowledge base of the ZSSP as well as the DM policy are adapted after each dialogue turn.

The second protocol, referred to as \textbf{RR} hereafter, directly adds the ZSSP learning actions to the dialogue manager RL policy, and therefore combines the two learning process into one single policy.

This variant of joint learning merges both policies in a single Q-learner. In that purpose the DM summary state vector has been augmented with a ZSSP-related dimension. Only one has been added so as to avoid dispersion of the policy which could arise from a larger increase in state size. 

This new dimension has been evaluated based on compound indices of the requirements for correction of the current annotation from the ZSSP. On a 3-point scale, three features have been used:
\begin{enumerate}
\item \textbf{confidence}: confidence score of the semantic parser in $[0,1]$
\item \textbf{fertility}: ratio of concepts on the utterance word length in $[0,1]$. As ZSSP tends to produce an over-segmentation of the incoming utterances with inserted concepts.
\item \textbf{rare}: binary presence of rare concepts in the annotation. Rare concepts are "help", "repeat", "restart", "reqalts", "reqmore", "ack" or "thankyou", and are generally wrongly annotated.
\end{enumerate}

From these features, an inform\_zssp feature is computed as:
\begin{enumerate}
\item[0] \textit{all clear}: confidence $>= 0.6$ and fertility $<= 0.2$ and rare $= 0$
\item[1] \textit{average condition}: confidence $>= 0.5$ and fertility $<= 0.7$ and rare $=0$ and (confidence $< 0.6$ or fertility $> 0.2$)
\item[2] \textit{alarming}: confidence $< 0.5$ or fertility $> 0.7$ or rare $= 1$
\end{enumerate}

At the same time the two ZSSP-annotation actions (Askconfirm and Askannotation, see previous section) are included in the list of summary actions dealt with by the dialogue policy and can be picked up by it. In such case the user is presented with the appropriate annotation window in the system's graphical interface and can correct the current annotation. Purely vocal interactions for this process are under study. Yet feasible it remains a challenging task which could introduce errors of its own, so it seemed more appropriate to evaluate the whole process first with a graphical interface and no input errors. Once done, the turn is updated (i.e. the annotation process has taken the place of the normal user audio response) and the dialogue is pursued. Even though it might be possible to have the policy learn it by itself we chose to inhibit two Ask actions in a row (they are tagged as impossible in the next turn). This two ZSSP-annotation actions have a specific social-feedback: the feedback $f_i$ uses the loss function described in Section~\ref{seq:zssp}, which is rescaled to obtain a score $\in[-1,1]$ :
$ f_i = (1.0 - l_i) * 2 - 1 $.

\section{Experimental Study}
\label{seq:exp}

\subsection{Task Description}

Experiments presented in this paper concern a chit-chat dialogue system framed in a goal-oriented dialogue task. In this context, users discuss with the system about an image (out of a small predefined set of 6), and they tried jointly to discover the message conveyed by the image, as described in~\cite{chaminade2017}. In order to use a goal-oriented system for such a task, the principle which has been followed was to construct, as the system's back-end, a database containing several hundreds of possible combinations of characteristics of the image, each associated with a hypothesis of the conveyed message. During its interaction with the system, the user by progressively giving elements from the image matching entities in the database will make the system select a small subset of possible entities from which it can pick both additional characteristics to inform the user with or ultimately a pre-defined message to give as a plausible explanation for the image's purpose. This would allow the user to speak rather freely about the image for several tens of seconds before arguing briefly about the message. No argumentation is possible from the system's side, it can only propose a canned message and the discussion is expected to last only around one minute at most. 

The task-dependent knowledge base used in the experiments is derived from INT task description~\cite{chaminade2017}, as well as from a generic dialogue information. The semantics of the task is represented by 16 different act types, 9 slots and 51 values. The lexical forms ($53$) used to model act types were manually elaborated.

\subsection{Results}

\begin{table*}[!ht]
\center
\begin{tabular}{ccccccc}
\textbf{Model} & \textbf{Train} & \textbf{Test} & \textbf{Success} & \textbf{Avg cum.} & \textbf{Sys. Underst.} & \textbf{Sys. Gener.} \\
& \textbf{(\#dial)} & \textbf{(\#dial)} & \textbf{(\%)} & \textbf{ Reward} & \textbf{ Rate} & \textbf{ Rate} \\
\hline
ZH           & 0   & 94 & 31 & -1.7 & 1.3 & 4.1 \\
BH - Train1 & 80  & 48 & 79 & 9.1 & 3.2 & 4.7 \\
BH - Train2 & 80  & 48 & 73 & 9.1 & 2.4 & 4.5  \\
BR - Train1 & 140 & 48 & 62 & 7.0 & 2.9 & 4.4  \\
BR - Train2 & 140 & 48 & 94 & 11.7 & 2.8 & 4.6 \\
RR - Train1 & 140 & 48 & 58 & 1.7 & 2.2 & 4.3  \\
RR - Train2 & 140 & 48 & 67 & 4.4 & 2.5 & 4.8  \\
\end{tabular}
\caption{Evaluation of the different configurations of on-line learning}
\label{table:evaluation}
\end{table*}

The evaluation of the two joint learning approaches is presented here. Two complementary systems are proposed in comparison: \textbf{ZH} is a baseline system without on-line learning using the initial ZSSP and a handcrafted dialogue manager policy, whereas the system \textbf{BH} combines the bandit on-line learning for ZSSP and the handcrafted dialogue manager policy.

For each system, two expert users communicated with the system to train two separate models. Then a group of (mostly) naive users tested each model. At the end of each session, the users were asked to give a rating on a scale of 0 (worst) to 5 (best) to the understanding and generation qualities of the system. The number of training dialogues, as well as the number of test sets for each configuration are given in Table~\ref{table:evaluation}.

\begin{figure*}[!ht]
  \begin{tabular}{ccc}

\includegraphics[width=0.5\textwidth]{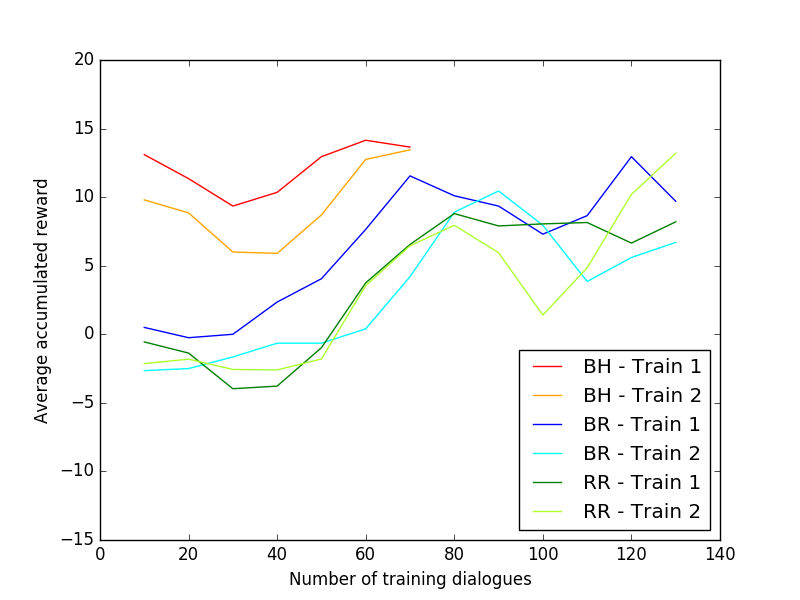}
		& 
  		\includegraphics[width=0.5\textwidth]{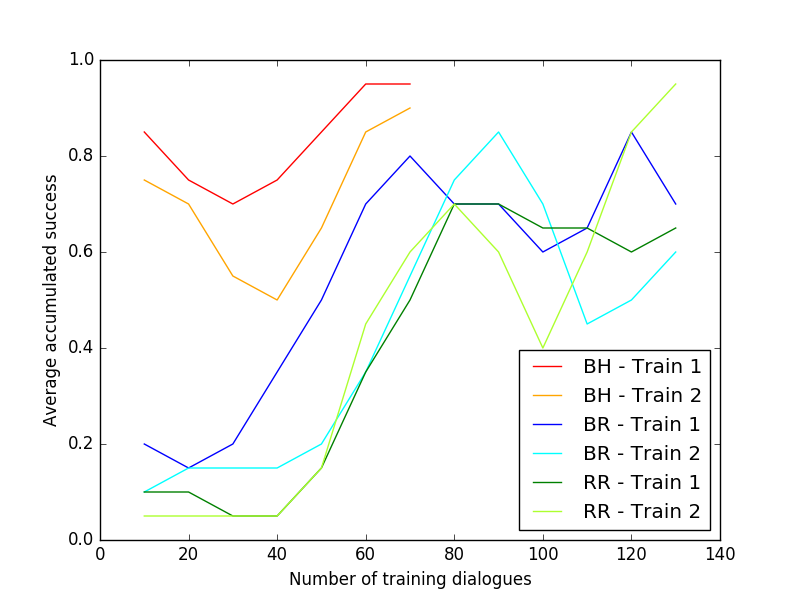}
   \end{tabular}
\vspace{-1pt}
\caption{\textbf{(a)} Average cumulated reward for different models
\textbf{(b)} Average cumulated success for different models}
\label{fig:results}
\end{figure*}

The dialogue success rate is calculated and the average cumulated reward is estimated for each protocol. Figure~\ref{fig:results} presents the learning curve for the BH, BR and RR systems. The curves are smoothed out by a 20-point sliding window with a 5-point shift. Regarding BH, we can observe that both trainings tend to initially have a high success rate while the expert users validate the baseline system. Then they started to diversify their inputs, thus quickly and temporarily degrading the results. Within fifty training dialogues, they achieved a higher success score. BR and RR start from scratch and initially have a very low success rate. At the beginning of the learning process, the expert is inclined to use simple dialogues to build and efficient dialogue manager policy, leading to a large increase of the success rate. Then, when the system starts to be usable, more sophisticated dialogue development are tested to teach more adaptability to the system. During the training, it drifts towards a decrease of the reward and success rates. The integration of the ZSSP learning actions to the dialogue manager RL policy presents some difficulties. Those actions are difficult to reward despite the social reward, since the most profitable ones usually happen during unsuccessful dialogues. Thus, they tend to quickly disappear during the learning and so to limit the ZSSP learning for RR systems. This can explain the lower scores of the RR systems during user tests discussed below.

For the BR model, it can be observed with Train2 that this protocol can reach very good performances (94\% success rate) with less than a hundred fifty training dialogues. Yet this performance seems to suffer of great variability, depending on the choices made by the expert. The experts may have the same tools at their disposal they still have a large margin of action in how they train their system: for instance they can decide to locally reward only the good actions, or reversely only the bad ones (or ideally, but more costly, both). They also have a large leeway on which kind of inputs they train the system with: either to stay very simple to ensure a steady learning curve or very realistic to try to reach immediately the interesting regions of the policy's state space. This is what can be observed in the differences between BR-Train1 (62\% succ rate) and BR-Train2 (94\%). However it is worth noticing that in both cases, the system performance is increasing with time, and so a system can be further improved to a certain level. Unfortunately, since these experiments are extremely expensive in time and require expert users to train the models, larger scale experiments are not affordable to statistically estimate the variance of performance of the resulting models over quite different experts applying different strategies.

The user trials of the two training trials for each protocol are given in Table~\ref{table:evaluation}. The results show that the different configurations of the system displays acceptable performance. The difference in performance between the ZH and the two BH models (+48\% absolute for Train1 and +42\% absolute for Train2) shows the impact of the ZSSP adaptation on the overall success of the conversation, along with a better understanding (1.3 for ZH vs. 3.2 for BH-Train1 and 2.4 for BH-Train2). The average cumulated reward rate on the test is directly correlated to the success rate and comes in confirmation of the previous observations. Also, due to a well-tuned template-based generation system, the system generation rate is high ($>4$) for all configurations.

The RR protocol offers smaller success rates compared to BH and BR (58\% for Train1 and 67\% for Train2 to be compared with 94\% for the best BR case). After analysing the training logs it seems to be related to the very low triggering level of the ZSSP learning actions after the exploration steps during RR w.r.t the use of the bandit (in BH and BR). To remedy this the policy state space should be modified to take a better account of the situations favorable to ZSSP actions, while preserving its capacities of discrimination for the dialogue actions. Anyhow this approach remains to be developed further and improved as it is based on a unique framework for joint learning, which simplifies the system elaboration from a programming point of view.  

\vspace*{-1.1ex}
\section{Conclusion}
\label{seq:concl}

After proposing methods to interactively train both semantic parsing and dialogue management on-line, this paper proposed and evaluated ways to combine them in a joint learning process. Experiments have been carried in real conditions and are therefore scarce. Yet it has been possible to show that joint learning can be operated, and that after only a hundred dialogues the performance of the various configurations tested were generally good enough compared to a handcrafted system.

Based on these results, we now investigate the possibility of merging the resulting policies between trials, so as to be able to pile up trainings coming from different users and save even more time to the system developers.

\bibliographystyle{IEEEbib}
\bibliography{refs,biblio_manuCSL}

\begin{thebibliography}{10}

\bibitem{Serban2016}
Iulian~V. Serban, Alessandro Sordoni, Yoshua Bengio, Aaron Courville, and
  Joelle Pineau,
\newblock ``Building end-to-end dialogue systems using generative hierarchical
  neural network models,''
\newblock in {\em Proceedings of the Thirtieth AAAI Conference on Artificial
  Intelligence}. 2016, AAAI'16, pp. 3776--3783, AAAI Press.

\bibitem{Wen2017}
Tsung-Hsien Wen, David Vandyke, Nikola Mrk{\v{s}}i{\'{c}}, Milica Gasic,
  Lina~M. Rojas~Barahona, Pei-Hao Su, Stefan Ultes, and Steve Young,
\newblock ``A network-based end-to-end trainable task-oriented dialogue
  system,''
\newblock in {\em Proceedings of the 15th Conference of the European Chapter of
  the Association for Computational Linguistics: Volume 1, Long Papers}. 2017,
  pp. 438--449, Association for Computational Linguistics.

\bibitem{Ferreira13b}
Emmanuel Ferreira and Fabrice Lef{\`e}vre,
\newblock ``Expert-based reward shaping and exploration scheme for boosting
  policy learning of dialogue management,''
\newblock in {\em ASRU}, 2013.

\bibitem{Gasic10}
M.~Ga\v{s}i\'{c}, F.~Jur\v{c}\'{\i}\v{c}ek, S.~Keizer, F.~Mairesse, B.~Thomson,
  K.~Yu, and S.~Young,
\newblock ``Gaussian processes for fast policy optimisation of pomdp-based
  dialogue managers,''
\newblock in {\em SIGDIAL}, 2010.

\bibitem{williams2013}
J.~Williams, A.~Raux, D.~Ramachandran, and A.~Black,
\newblock ``The dialog state tracking challenge,''
\newblock in {\em SIGDIAL}, 2013.

\bibitem{Casanueva2017}
I{\~{n}}igo Casanueva, Pawe{\l} Budzianowski, Pei-Hao Su, Nikola
  Mrk{\v{s}}i{\'{c}}, Tsung-Hsien Wen, Stefan Ultes, Lina Rojas-Barahona, Steve
  Young, and Milica Ga{\v{s}}i{\'{c}},
\newblock ``{A Benchmarking Environment for Reinforcement Learning Based Task
  Oriented Dialogue Management},''
\newblock {\em arxiv.org}, nov 2017.

\bibitem{Young10}
S.~Young, M.~Ga\v{s}i\'{c}, S.~Keizer, F.~Mairesse, J.~Schatzmann, B.~Thomson,
  and K.~Yu,
\newblock ``The hidden information state model: A practical framework for
  pomdp-based spoken dialogue management,''
\newblock {\em Computer Speech and Language}, vol. 24, no. 2, pp. 150--174,
  2010.

\bibitem{Daubigney12}
L.~Daubigney, M.~Geist, S.~Chandramohan, and O.~Pietquin,
\newblock ``A comprehensive reinforcement learning framework for dialogue
  management optimization,''
\newblock {\em Selected Topics in Signal Processing}, vol. 6, no. 8, pp.
  891--902, 2012.

\bibitem{FerreiraCSL}
Emmanuel Ferreira and Fabrice Lef{\`{e}}vre,
\newblock ``Reinforcement-learning based dialogue system for human-robot
  interactions with socially-inspired rewards,''
\newblock {\em Computer Speech {\&} Language}, vol. 34, no. 1, pp. 256--274,
  2015.

\bibitem{ferreira2015}
E~Ferreira, B~Jabaian, and F~Lef{\`e}vre,
\newblock ``Online adaptative zero-shot learning spoken language understanding
  using word-embedding,''
\newblock in {\em ICASSP}, 2015.

\bibitem{Ferreira2016}
Emmanuel Ferreira, Alexandre Reiffers-Masson, Bassam Jabaian, and Fabrice
  Lef\`evre,
\newblock ``Adversarial bandit for online interactive active learning of
  zero-shot spoken language understanding,''
\newblock in {\em Proceedings of ICASSP}, 2016.

\bibitem{Riou2017}
Matthieu Riou, Bassam Jabaian, St{\'{e}}phane Huet, and Fabrice Lef{\`{e}}vre,
\newblock ``Online adaptation of an attention-based neural network for natural
  language generation,''
\newblock in {\em Proceedings of INTERSPEECH}, 2017.

\bibitem{hahn2010}
S.~Hahn, M.~Dinarelli, C.~Raymond, F.~Lef{\`e}vre, P.~Lehnen, R.~{De Mori},
  A.~Moschitti, H.~Ney, and G.~Riccardi,
\newblock ``{Comparing stochastic approaches to spoken language understanding
  in multiple languages},''
\newblock {\em IEEE TASLP}, vol. 19, no. 6, pp. 1569--1583, 2010.

\bibitem{lefevre2007}
F.~Lef{\`e}vre,
\newblock ``{Dynamic Bayesian networks and discriminative classifiers for
  multi-stage semantic interpretation},''
\newblock in {\em {ICASSP}}, 2007.

\bibitem{deoras2013}
A.~Deoras and R.~Sarikaya,
\newblock ``Deep belief network based semantic taggers for spoken language
  understanding,''
\newblock in {\em INTERSPEECH}, 2013.

\bibitem{dauphin2014}
Y.~Dauphin, G.~Tur, D.~Hakkani-Tur, and L.~Heck,
\newblock ``Zero-shot learning and clustering for semantic utterance
  classification,''
\newblock {\em arXiv preprint arXiv:1401.0509}, 2014.

\bibitem{mikolov2013}
T.~Mikolov, K.~Chen, G.~Corrado, and J.~Dean,
\newblock ``Efficient estimation of word representations in vector space,''
\newblock {\em arXiv preprint arXiv:1301.3781}, 2013.

\bibitem{Geist10}
M.~Geist and O.~Pietquin,
\newblock ``Kalman temporal differences,''
\newblock {\em Artificial Intelligence Research}, vol. 39, no. 1, pp. 483--532,
  Sept. 2010.

\bibitem{Ng99}
A.~Ng, D.~Harada, and S.~Russell,
\newblock ``Policy invariance under reward transformations: Theory and
  application to reward shaping,''
\newblock in {\em ICML}, 1999.

\bibitem{bian2014}
J.~Bian, B.~Gao, and T.~Liu,
\newblock ``Knowledge-powered deep learning for word embedding,''
\newblock in {\em ECML}, 2014.

\bibitem{geist2010managing}
Matthieu Geist and Olivier Pietquin,
\newblock ``Managing uncertainty within value function approximation in
  reinforcement learning,''
\newblock in {\em Active Learning and Experimental Design workshop (collocated
  with AISTATS 2010), Sardinia, Italy}, 2010, vol.~92.

\bibitem{chaminade2017}
Thierry Chaminade,
\newblock ``{An experimental approach to study the physiology of natural social
  interactions},''
\newblock {\em {Interaction Studies}}, vol. 18, no. 2, pp. 254--276, 2017.

\end{thebibliography}

\end{document}